\def\year{2022}\relax
\documentclass[letterpaper]{article} 
\usepackage{aaai22}  
\usepackage{times}  
\usepackage{helvet}  
\usepackage{courier}  
\usepackage[hyphens]{url}  
\usepackage{graphicx} 
\usepackage{natbib}  
\usepackage{caption} 
\DeclareCaptionStyle{ruled}{labelfont=normalfont,labelsep=colon,strut=off} 
\usepackage{algorithm}
\usepackage{algorithmic}
\usepackage{newfloat}
\usepackage{listings}
\usepackage{multirow}
\usepackage{tabularx}
\usepackage{hyperref}
\usepackage[dvipsnames]{xcolor}

\floatstyle{ruled}
\newfloat{listing}{tb}{lst}{}
\floatname{listing}{Listing}
\usepackage{graphicx}
\usepackage{adjustbox}
\usepackage{tabularx}
\usepackage{comment}
\usepackage{amsmath}
\usepackage{mwe}
\usepackage{caption}
\usepackage{booktabs}

\usepackage{color}
\usepackage[T1]{fontenc}
\usepackage[utf8]{inputenc}

\usepackage{microtype}

\title{Self-Supervised Knowledge Assimilation for Expert-Layman Text Style Transfer}

\author{
    Wenda Xu, Michael Saxon, Misha Sra, William Yang Wang
}
\affiliations{
    University of California, Santa Barbara\\

    Department of Computer Science\\
    Santa Barbara, California, USA\\
    \texttt{\{wendaxu,saxon\}@ucsb.edu \{sra,william\}@cs.ucsb.edu}
}

\begin{document}
\maketitle
\begin{abstract}
Expert-layman text style transfer technologies 
have the potential to improve communication between members of scientific communities and the general public. High-quality information produced by experts is often filled with difficult jargon laypeople struggle to understand. This is a particularly notable issue in the medical domain, 
where layman are often confused by medical text online.
At present, two bottlenecks interfere with the goal of building high-quality medical expert-layman style transfer systems: a dearth of pretrained  medical-domain language models spanning both expert and layman terminologies and a lack of parallel corpora for training the transfer task itself. To mitigate the first issue, we propose a novel language model (LM) pretraining task, 
\textit{Knowledge Base Assimilation}, 
to synthesize pretraining data from the edges of a graph of expert- and layman-style medical terminology terms into an LM during self-supervised learning.
To mitigate the second issue, we build a large-scale parallel corpus in the medical expert-layman domain using a margin-based criterion.
Our experiments show that transformer-based models pretrained on knowledge base assimilation and other well-established pretraining tasks fine-tuning on our new parallel corpus leads to considerable improvement against expert-layman transfer benchmarks, gaining an average relative improvement of our human evaluation, the Overall Success Rate (OSR), by 106\%. We release our code and parallel corpus for future research \footnote{Code available at \href{https://github.com/xu1998hz/SSL_KBA_Expert_Layman_Style_Transfer}{\color{Blue}{\texttt{\url{https://github.com/xu1998hz/SSL_KBA_Expert_Layman_Style_Transfer}}}}.}.


\end{abstract}

\section{Introduction}
Incompatible knowledge backgrounds between experts and laymen cause communication difficulties \cite{Jerit2009UnderstandingTK}. 
These difficulties are particularly problematic in the medical domain when patients attempt to self-diagnose their ailments online \cite{white2010web}. Their search terms might be too vague, leading them to self-misdiagnose, followed by unnecessary treatment or tests, and potentially worse outcomes \cite{Au462}. 
Even if they find high-quality, correct online medical resources that match their condition, the incomprehensible medical jargon within can be confusing and frustrating \cite{benigeri2003shortcomings}. 
Misunderstandings from online medical information seeking have been shown to lead to increased health anxiety \cite{white2009experiences}.
Expert-layman text style transfer technologies offer a potential method to resolve these problems. Accurate  layman-to-expert style conversion of vague searches into precise terminology could improve the quality of retrieved documents. In turn, high-quality expert-to-layman translation of the retrieved documents would lead to more comfort, better understanding, and hopefully better overall outcomes. 

Text style transfer is the task of transforming a passage of text from a source style (e.g., expert medical language) to a target style (e.g., layman language) while preserving the underlying meaning \cite{jin2021deep}.
Prior work has demonstrated impressive style transfer results across a variety of attributes, including sentiment  \cite{li-etal-2018-delete, dai2019style}, formality \cite{rao2018dear}, and politeness \cite{inproceedings_2}. 
However, the aforementioned style-transfer tasks are fundamentally surface-level transformations along fairly content-agnostic dimensions. Expert-layman style transfer is different in that it requires \textit{domain-specific terminological correspondence knowledge}. For a given domain, the model must contain sets of mappings between specific expert and layman-style expressions for phenomena (e.g., ``renal'' and ``relating to the kidneys''), and a system intended for one domain has no guaranteed applicability to another (e.g., linguistics to medicine).



In this paper, we tackle two core hurdles to building high-quality medical expert-layman style transfer systems: a lack of pretrained sequence-to-sequence language models containing medical domain-specific terminological correspondence knowledge, and a lack of parallel expert-layman medical corpora for fine-tuning. 
For the first hurdle, we introduce a novel language model (LM) pretraining task, \textbf{knowledge base assimilation} (KBA) to explicitly provide the LM with a learning signal across expert-layman phrasal realizations of concepts. We further augment our KBA training with the previously proposed \textit{Mask}, \textit{Switch}, and \textit{Delete} self-supervised pretraining tasks \cite{devlin2019bert, lample2018unsupervised, lewis2019bart} 
to build a robust medical LM containing terminological correspondence knowledge. To the best of our knowledge, this is the first work to investigate self-supervised representation learning in expert-layman text style transfer. To tackle the second hurdle, we produce a high-quality parallel extension of the non-parallel MSD medical text dataset \cite{cao-etal-2020-expertise} 
containing 11,512 expert- and layman-style medical sentences using a margin-based data mining criterion \cite{schwenk-2018-filtering}. To the best of our knowledge, this is the first work to utilize supervised learning on a large data mined parallel corpus of expert-laymen sentences.

\begin{figure*}[t!]
    \small
    \centering
    \includegraphics[width=\linewidth]{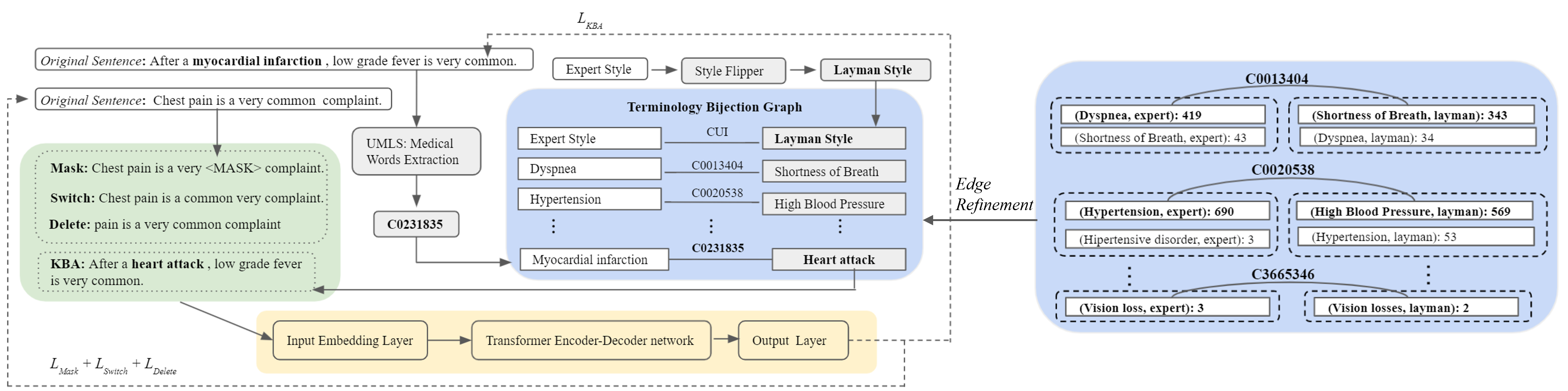}
    \caption[width=\linewidth]{Multi-Task Self-Supervised Learning integrates KBA with \textit{Mask, Switch, Delete} tasks. Four losses are optimized concurrently. The parameters of input embedding layer, encoder-decoder network and output layer are shared between pretraining and fine-tuning. On the right side, we demonstrate the construction of the Terminology Bijective Graph through edge refinement.}
    \label{fig:selfsupervised}
\end{figure*}

\section{Proposed Approach}



A \textit{knowledge domain} $D$ contains many \textit{concepts} $C_i$, which are sets of sentences $S_{i,j}$ of equivalent meaning. Sentences can be labeled with attributes, including whether they belong to the ``expert'' or ``layman'' style. We define the task of expert-layman style transfer as follows: given a sentence $S\in C_i$ with either the expert- or layman-style, generate a sentence $S'$ in the other style that is also a member of $C_i$. This is not a simple task, as it requires the model to be aware of the underlying medical concept that links semantically equivalent but lexically unique medical phrases together. 

To tackle this problem, we first pretrain a transformer language model on an ensemble of tasks, including our novel \textbf{knowledge base assimilation} (KBA) task and the previously demonstrated \textit{Mask, Delete} and \textit{Switch} \textbf{self-supervised learning} (SSL) tasks, to simultaneously model medical language in general while also capturing how specific concepts are phrased in each style. Then, we fine-tune this language model on \textbf{a new corpus of parallel expert-layman medical sentences} we extract using a margin-based criterion from the unaligned MSD dataset.

\begin{table}[b!]
  \label{dataset_table}
  \centering
  \small
\resizebox{0.45\textwidth}{!}{%
  \begin{tabular}{llll}
    \toprule
    Number of... & Expert & Layman & Ratio \\
    \midrule
    MSD Training Sentences & 130,349 & 114,674 & 0.88 \\
    MSD Test Sentences & 657& 657& 1\\ \midrule
    KBA Generated Pretraining Sentences & 40892 & 31083 & 0.75 \\
    Data-Mined Parallel Sentences & 11512 & 11512 & 1 \\
    Medical Concepts (CUI codes) & 10810 & 10810 & 1 \\
    Terminology Bijective Graph Triples & 1124 & 1124 & 1 \\
    \bottomrule
  \end{tabular} 
 }
  \caption{MSD dataset statistics, and quantities for our derived KBA pretraining and fine-tuning data. }
  \label{tab:msd}
\end{table}

\subsection{Dataset}
We evaluate our proposed method and current SOTA models using the MSD dataset \cite{cao-etal-2020-expertise}. To our knowledge, it is the only available dataset for the task of  medical expert-layman text style transfer. MSD contains 245k medical training sentences which are each labeled with either the ``expert'' or ``layman'' style. Additionally, it contains a test set of 675 expert-layman sentence pairs of equivalent meaning. We extend the training set by producing 11,512 sentence pairs using a margin-based criterion \cite{schwenk-2018-filtering}. There are 10810 medical concepts which are used by both expert and layman sentences. We use our edge refinement to create 1124 triples in our Terminology Bijective Graph and build 40,892 expert and 31,083 layman pseudo training sentences for KBA task. Specific statistics are listed in Table \ref{tab:msd}.

\section{Pretraining Strategy}

We use the standard transformer encoder-decoder \cite{vaswani2017attention} with 4 layers, 4 attention heads, and a hidden size of 256 as our language model. We perform a multi-task pretraining procedure where 
we train a single shared feedforward encoder-decoder framework across KBA and the three SSL tasks, 
\textit{Mask, Switch} and \textit{Delete}. For the KBA task, we construct 71,975 training sentences from MSD training set using a \textit{Terminology Bijective Graph} described below. We construct training data for \textit{Mask, Switch, Delete} tasks by separately applying their respective noise functions for each sentence in the MSD training dataset, as depicted in Figure 1. We optimize on all four pretraining tasks concurrently, by minimizing negative log-likelihoods: 

\begin{equation}
\mathcal{L}_\text{total} = \mathcal{L}_\text{KBA} +\mathcal{L}_\text{Mask} + \mathcal{L}_\text{Switch} + \mathcal{L}_\text{Delete}
\label{eq:L}
\end{equation}

Each mini-batch contains even distribution of each task's pretraining data. 
Details provided below.

\subsection{Knowledge Base Assimilation}\label{sec:replace} Knowledge Base Assimilation (KBA) resembles sequence-to-sequence knowledge distillation, except it utilizes a KB 
rather than a teacher model during training. In particular, we generate a sentence $S'$ from a sentence $S$ in either expert or layman style, where each term in $S$ (a node in the KB) is replaced by a term with the same meaning but in the opposite style (a node opposite the original term along an edge in the KB) shown in Figure \ref{fig:selfsupervised}. The training task then becomes reconstructing $S$ from $S'$ by replacing all the terms, thereby training the LM to model edges in the Terminology Bijective Graph.

In particular, given a source sentence denoted by $S=\{w_1, m_2, w_3, w_4, m_5, ...w_n\}$, where $w_i$ denotes a non-medical word and $m_j$ denotes a medical phrase in the source style. The target style medical phrase which has the same meaning as $m_j$ is denoted by $m'_j$. Both $m_j$ and $m'_j$ are connected by an edge in the Terminology Bijective Graph. The input of the KBA task is the sentences with the replaced medical phrases in the target style, $S' = \{w_1, m'_2, w_3, w_4, m'_5, ...w_n\}$. The model is required to reconstruct the original sentence from the replaced input sentence. The purpose of the KBA is to enable the model to pick out medical phrases which are misaligned with the sentence style and learn the mapping of concept pairs with identical meaning. We illustrate this process in Figure \ref{fig:selfsupervised}.

\subsection{Terminology Bijective Graph}
\label{sec:word_map_table}
To perform KBA for expert-layman transfer in the medical domain, we require a knowledge base of expert term-layman term relation correspondence edges. To achieve this, we build a child knowledge base of the Unified Medical Language System (UMLS) \cite{journals/nar/Bodenreider04} containing terms that appear in the aforementioned MSD dataset \cite{cao-etal-2020-expertise}.

The UMLS is a standardized knowledge base maintained by the United States National Library of Medicine, containing a collection of Concept Unique Identifier (CUI) codes and corresponding descriptions. CUI codes provide fixed reference to ``medical concepts'' that are invariant to language or style (i.e., expert vs. layman). For example, CUI 'C0013404' corresponds to both `dyspnea' and `shortness of breath.' \citet{cao2018neural} match every medical term in the MSD dataset to its CUI code using QuickUMLS \cite{Soldaini2016QuickUMLSAF}. 

Knowing the CUI codes that appear in MSD a-priori, we construct a child knowledge base of UMLS, the \textit{Terminology Bijective Graph}, exclusively containing pairs of terms with shared CUI in the expert and layman styles connected by a bidirectional ``is in the other style'' relation. To do this, we first collect 10,810 CUI codes which are both found in expert and layman sentences of MSD dataset. Based on those CUI codes, we form 10,810 medical phrasing subsets in each of expert and layman style. However, some phrasings for a CUI appear in both the expert- and layman-labeled medical phrasing subsets. Furthermore, sometimes two phrasings shared the same CUI code in opposite styles are just grammatical variations. We indicate both cases in the right side of Figure \ref{fig:selfsupervised}. To select the best candidate for each style in these cases, we apply a set of heuristics for \textit{edge refinement}.


We first select our expert and layman terms from the candidates. For each CUI, we select one phrase to be the expert term and the other to be layman by a simple majority vote of the MSD style label for the sentences they appear in. For each medical phrasing subset of expert and layman style, the (term, style label) pair with the highest frequency is selected.
 Thus, each CUI code provides a connection in the Terminology Bijective Graph giving the correspondence between the most frequently used expert and layman phrasing of the underlying concept. Then, we apply Levenshtein distance \cite{10.5555/1822502} with a threshold ($d=4$) to exclude candidate phrasing pairs which are simply grammatical variations. This process is shown in the right side of Figure \ref{fig:selfsupervised}.

After the process of edge refinement we are left with a high-quality Terminology Bijective Graph containing 1,124 expert-layman edges with which we perform KBA. Although domain-specific terminological correspondence only requires one relation in the Terminology Bijective Graph, future works can extend KBA to assimilate more complex knowledge graph structures into SSL to novel domains.

\subsection{Self-Supervised Learning Tasks}\label{sec:selfsup}
In Figure \ref{fig:selfsupervised}, we further augment the KBA edge modeling task with previously demonstrated self-supervised learning (SSL) tasks \textit{Mask, Switch}, and \textit{Delete} on the full MSD training corpus to build a more robust model of medical language. All three are sequence-to-sequence denoising autoencoder tasks, where an original sentence $S$ is the reconstruction target given a perturbed input sentence $S'$. Details for each SSL task are provided below:


\subsubsection{Mask}
The \textit{Mask} task follows the BERT \cite{devlin2019bert} masking scheme. 15\% of the word tokens in $S$ are randomly replaced by <MASK> tokens to produce $S'$.

\subsubsection{Switch}
The \textit{Switch} task generates $S'$ by shuffling the word order of a sentence. For each sentence $S$, we first select $15\%$ of words in $S$ at random to be shuffled. Then, the selected words are randomly reordered amongst themselves while preserving the order of the unselected words to generate $S'$.
This is similar to \citet{lample2018unsupervised}'s noise function.

\subsubsection{Delete}
In the \textit{Delete} task, we randomly delete 15\% of the word tokens.
In contrast to the Mask Task, the \textit{Delete} task requires the model to learn not only the contextual information for the deleted tokens but also learn the possible positions to insert words. 
This is similar to the token deletion pretraining task for BART \cite{lewis2019bart}.



\begin{figure}
    \centering
    \small
    \includegraphics[width=\linewidth]{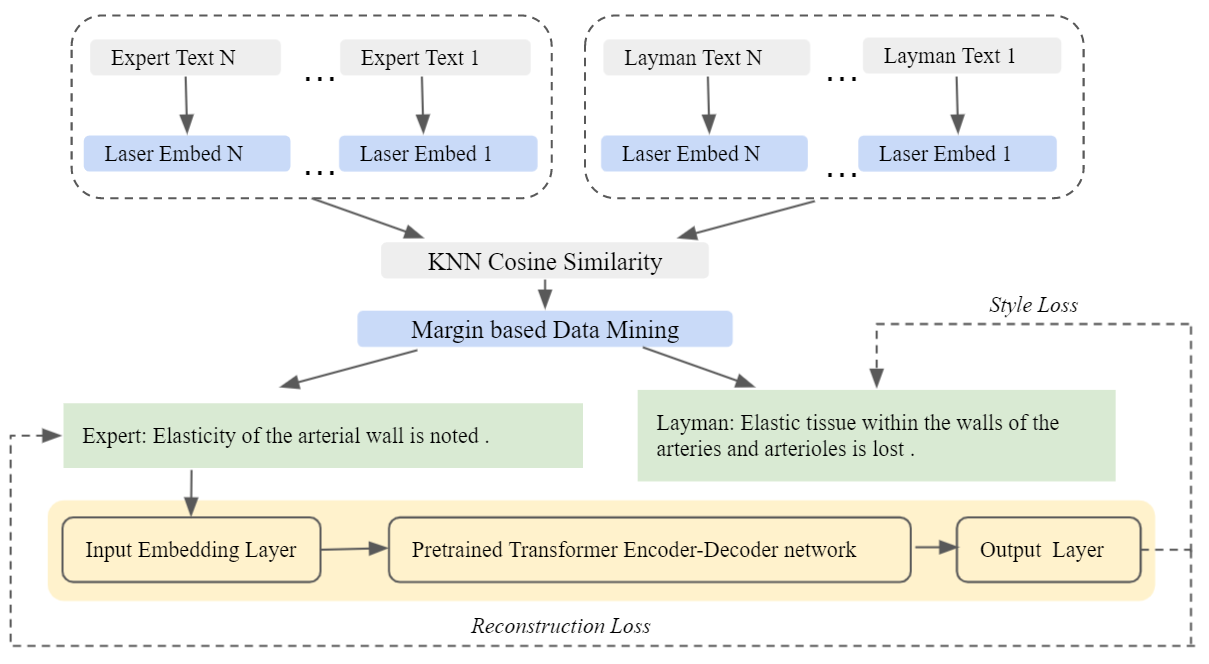}
    \caption{Data mining + Style Transfer Fine-tuning pipeline: Style and Reconstruction losses are optimized concurrently.}
    \label{fig:laser-pipeline}
\end{figure}

\section{Fine-tuning for Style Transfer}

With our pretrained medical domain encoder-decoder language model, we turn to supervised learning to train an expert-layman medical style transfer model. This requires a parallel corpus of medical sentences that share the same meaning, $(S_{i,\text{layman}},S_{i,\text{expert}})\in C_i$. We collect such a corpus by pairing sentences in the MSD dataset using a margin-based criterion. 

Since the transformer already has considerable in-domain modeling capabilities from pretraining, fine-tuning converges very fast. 
During the fine-tuning stage, the Transformer is modeled by two losses: self-reconstruction and style transfer loss. We use a similar encoding strategy as Style Transformer \cite{dai2019style}, in which we encode a style embedding into the input. 
Two losses are optimized concurrently, by minimizing two negative log-likelihoods, Equation \ref{eq:L2}. Therefore, the model will preserve content while rewriting sentences into the target style, shown in Figure \ref{fig:laser-pipeline}.


\begin{equation}
\mathcal{L}_\text{total} = \mathcal{L}_\text{Self-Reconstruction} + \mathcal{L}_\text{Style} 
\label{eq:L2}
\end{equation}

\subsection{Data Mining with Margin Criterion}
To collect training data for fine-tuning, we extract 11,512 paired sentences from 245k sentences (approximately 10\%) of the MSD training set \cite{cao-etal-2020-expertise} using margin-based criterion \cite{schwenk-2018-filtering}. This new parallel corpus is used for fine-tuning of the Transformer model, in Figure \ref{fig:laser-pipeline}. 

We first extract LASER embeddings \cite{DBLP:journals/corr/abs-1812-10464} 
of all sentences in both the expert and layman sets.
The margin is defined as the ratio of cosine similarity of two sentence embeddings and the average cosine similarities of k-nearest neighbors in both forward and backward directions. $x$ stands for one sentence in the source style set and $y$ stands for one sentence in the target style set. $N_k(x)$ stands for $k$ unique nearest neighbors of $x$ in the target style set. Similarly for $y$, the $k$ nearest neighbors are $N_k(y)$.

\begingroup
\begin{equation} 
M(x, y) = \frac{\cos(x, y)}{\sum_{z \in N_k{(x)}} 
\frac{\cos(x, z)}{2k} + \sum_{z \in N_k{(y)}} 
\frac{\cos(y, z)}{2k}}
\label{eq:1}
\end{equation} 
\endgroup

We use the ``\textit{max-strategy}'' from \cite{schwenk-2018-filtering} to calculate margin in both directions (expert to layman and layman to expert) for all sentences in both style sets. That allows us to build candidate pairs for both directions (expert$-$to$-$layman and layman$-$to$-$expert). Any sentence can occur, at most, once in the candidate pairs. Therefore, other candidate pairs of that sentence with smaller margin values will be excluded. We use a threshold on the margin score to select candidate sentence pairs that are mutual translations of each other. Discussion on mutual translations can be found in \cite{schwenk-2018-filtering}. Margin criteria for our model is set as $k=4$ and $threshold = 1.06$. Parallel corpus generation took 7.5 hours on a single Titan 1080 Ti GPU.

\section{Experiments}

We assess the performance of our strategy by pretraining our transformer encoder-decoder LM in four different conditions: 
\begin{enumerate}
    \item \textbf{Basic}, where the transformer model recieves only self-reconstruction loss in pretraining.
    \item \textbf{SSL Only}, where only the three SSL tasks are used.
    \item \textbf{KBA Only}, where only KBA pretraining is used.
    \item \textbf{KBA+SSL}, where the model is simultaneously pretrained using KBA and the three SSL tasks.
\end{enumerate}
For all four conditions, we fine-tune the resulting LM on the expert-layman transfer task using the paired dataset. Furthermore, we compare our model with three prior baselines.

\subsection{Baseline Models} 
Following \citet{cao-etal-2020-expertise}, we choose baseline models that are commonly used and have publicly available code. 
\begin{enumerate}
    \item \textbf{DAR} \cite{li-etal-2018-delete} reconstructs the input sequence after source-target word replacement using edit distance.
    \item \textbf{Style Transformer} \cite{dai2019style} uses cyclic reconstruction to preserve content while doing style transfer.
    \item \textbf{Controlled Generation} \cite{hu2018controlled} uses a variational encoder to reconstruct the content representation and an attribute discriminator to build the style vector.
\end{enumerate}
None of these baseline models specifically deal with style transfer in the medical domain.

\subsection{Training Details}
We use the standard training settings for all models with Adam optimizer \cite{DBLP:journals/corr/KingmaB14} and early stops applied. Max sequence length, learning rate and drop out rate are set to $100, 1e-4$ and $0.5$ respectively. 
The three baselines, being unsupervised, are only trained on the non-parallel MSD corpus; they cannot be fine-tuned on our parallel corpus without modification. 

Our model architecture follows \citet{dai2019style}, with 4 layers, 4 attention heads per layer, and hidden size $256$. We add one style token into the input sequence with $256$ hidden units after the embedding layer. The positional encoding is applied to the entire input sequence except style embedding. For different SSL task combinations, the pretraining took 6 hours on average and fine-tuning took 1.5 hours on a single Titan 1080 Ti GPU. We use clinical-BERT's \cite{huang2020clinicalbert} tokenization for all models.

Finally, we augment our expected ``best'' condition of \textbf{KBA + SSL} pretraining by making \textbf{KBA + SSL Large}, identical to the other transformer models but for a hidden size of 512. We pretrain and fine-tune this model identically to the others.

\subsection{Human Evaluation}
We hired crowdworkers on Amazon Mechanical Turk\footnote{Each crowdworker, from the English-speaking locales of \{US, CA, UK, AU, NZ, IE\} was paid \$0.80 per task and averaged 5.5 minutes of completion time with an average compensation of \$8.73/hr.} (AMT) to rate the output of all systems.
We collected a random subset of 500 MSD test set sentences to evaluate the performance of all our models. For each source sentence and its style-transferred output, workers were asked to rate the output on three aspects: \textbf{content preservation} (the extent to which the two sentences match), \textbf{style transfer strength} (extent to which the desired change in style takes place), and \textbf{grammar fluency}. Crowdworkers answer questions on each aspect of a set of translations on a 5 point Likert scale. 

To ensure the layman crowdworkers understood both sentences, we include supplementary medical definitions for all medical terms in each sentence. Crowd-workers were able to access those definitions with a mouse-over of the underlined medical words in the interface (See Appendix Figure \ref{fig:mturk} and Appendix ``Implementation of Human Evaluation Interface'').

Due to the knowledge gap between expert and layman sentences, the understanding comparison between the transferred sentence and the source sentence is the most direct way to assess the strength of expert-layman text style transfer. The comparison is quantified by the number of times that crowd-workers had to check supplementary definitions of medical words in the sentence. Fewer checks in the transferred sentence imply an easier-to-understand sentence compared to the source sentence, and vice versa. In the expert to layman (E2L) direction, higher understanding score means the transferred sentence is easier for laymen annotators to understand, while in the direction of layman to expert (L2E), a higher score indicates the transferred sentence is harder to understand.

Following 
\citet{li-etal-2018-delete}, we report six success rates. We consider a transferred sentence ``successful'' in one evaluation criteria if it is rated 4 or 5 by AMT workers. 

\paragraph{Content Success Rate}(CSR)---the percentage of sentences that receive 4 or 5 rating in the content criterion.

\paragraph{Understanding Success Rate}(USR)---the percentage of sentences that receive 4 or 5 rating for ``understand.''

\paragraph{Grammar Success Rate} (GSR)---the percentage of sentences that receive 4 or 5 rating for the grammar criterion.

\paragraph{Understanding + Content Success Rate} (UCSR)---the percentage of sentences that receive 4 or 5 rating for both content and ``understand'' criteria.

\paragraph{Understanding + Grammar Success Rate} (UGSR)---the percentage of sentences that receive 4 or 5 rating for both grammar and ``understand'' criteria.

\paragraph{Overall Success Rate} (OSR)---the percentage of sentences that receive 4 or 5 rating in all three criteria.\\

We use CSR, USR and GSR to directly assess the model's performance in content preservation, style transfer strength and grammar fluency respectively. We further define a concept, \textit{effective style transfer}: An effective style transfer happens only when model can preserve the sentence meaning or fluency during Style Transfer. We include UCSR, UGSR, OSR to indicate the percentage of sentences that can achieve effective style transfer and OSR can also reflect the overall performance of models on three criteria.


\subsection{Automated Evaluation}

Following previous work by \cite{dai2019style, cao-etal-2020-expertise}, we compute three automatic evaluation metrics (see Table \ref{tab:autoeval}). 
We train a style classifier on the MSD training set using FastText \cite{joulin2016fasttextzip} to estimate the style accuracy of the transferred sentence. The \textbf{style classifier score} indicates the percentage of the transferred sentences that labeled as corresponding style. We also use NLTK \cite{bird2009natural} to calculate \textbf{4-gram BLEU} \cite{papineni-etal-2002-bleu} scores between the transferred sentence and the original sentence. We use KenLM \cite{Heafield11kenlm:faster} to train a 5-gram language model on the MSD training set. We use \textbf{PPL} to measure the fluency of the transferred sentence.

\section{Results}
From our human evaluation results in Table \ref{tab:ssl}, our model trained on the SSL task only achieves the highest CSR score. Our KBA+SSL and KBA+SSL Large variants achieve the highest USR, indicating their most progressive style transfer strengths. Although, ControlledGen (CtrlGen) \cite{hu2018controlled} achieves the highest GSR, it mostly copies input to the output with limited style transfer, as indicated by its low USR and OSR. All baseline models only achieve limited effective style transfer, as indicated by their low UCSR, UGSR and OSR. In contrast, our KBA+SSL Large has relative improvements over the best performing baseline model DeleteAndRetrieve (DAR) \cite{li-etal-2018-delete} by 39\% in UCSR, 59\% in UGSR and 75.6\% in OSR.
To better understand the characteristics of the models, we provide a case study of an expert-to-layman input example and a layman-to-expert example in Table \ref{tab:casestudy}. 

\begin{table}
  \label{results_table}
  \small
  \resizebox{0.48\textwidth}{!}{%
  \begin{tabular}{lllllll}
    \toprule
    \multicolumn{1}{c}{} & \multicolumn{6}{c}{Human Evaluation} \\
    \cmidrule(r){2-7}
    Model & CSR & USR & GSR & UCSR & UGSR & OSR \\
    \midrule
    Style Tr \cite{dai2019style} & 0.703 & 0.281 & 0.615 & 0.176 & 0.058 & 0.113 \\
    DAR \cite{li-etal-2018-delete} & 0.695 & 0.321 & 0.472 & 0.231 & 0.148 & 0.123 \\
    CtrlGen \cite{hu2018controlled} & 0.852 & 0.195 & \textbf{0.739} & 0.103 & 0.147 & 0.086 \\
     \midrule
    Basic Tr. & 0.683 & 0.301 & 0.465 & 0.161 & 0.088 & 0.068 \\
    KBA Pretraining Only & 0.733 & 0.315 & 0.600 & 0.194 & 0.131 & 0.113\\
    SSL Pretraining Only & \textbf{0.870} & 0.303 & 0.732 & 0.257 & 0.215 & 0.200\\
    KBA + SSL Pretraining & 0.825 & \textbf{0.373} & 0.652 & 0.301 & 0.224 & 0.200 \\
    KBA + SSL Large (512) & 0.860 & \textbf{0.373} & 0.666 & \textbf{0.320} & \textbf{0.235} & \textbf{0.216} \\
    \bottomrule
  \end{tabular}
    }
    \caption{Human Evaluation Table: Our supervised models are below the middle line, the unsupervised baselines are above. 
    }
  \label{tab:ssl}
\end{table}

\subsubsection{Effects of KBA and SSL Tasks} 
Table \ref{tab:ssl} shows that even the \textbf{basic transformer} model without specialized SSL or KBA pretraining achieves competitive USR compared to the baseline models on fine-tuning alone. However, its outputs tend to lose the original sentence meaning or fluency while it adapts to our parallel corpus, indicated by its low CSR and GSR. Moreover, its effective style transfer is limited, indicated by its low UCSR, UGSR and OSR. 

Adding \textbf{KBA} training (KBA only condition) improves USR by learning terminological mappings between the styles. Surprisingly, as the model learns to reconstruct sentences from the replaced target style medical words both content and grammar scores are improved. Those improvements over all three criteria leads to the enhance of effective style transfer, indicating by UCSR, UGSR and OSR. 

Using the auxiliary learning of the \textbf{SSL} tasks significantly improves CSR and GSR. This finding verifies our assumptions that context-aware learning gives rise to better apparent content understanding and syntactic awareness. Interestingly, this multitask context-aware learning also demonstrates robust performance in effective style transfer, leading to steep increases of all UCSR, UGSR and OSR. 

By adding context-aware learning into KBA (KBA+SSL condition), we observe improvements across all criteria, demonstrating the importance of context-aware learning to the final pretraining scheme. By adding KBA to SSL, we observe consistent improvements of USR, UCSR and UGSR. This finding suggests that the shared representation of context-aware understandings and terminology mappings can improve style transfer strength and this improvement is ``effective'' in considering content and fluency. However, since the learning of KBA is directly enforced through our Terminology Bijective Graph, KBA-generated representations only have limited sentence context, leading to drops in CSR and GSR in the composed setting, compared to the more context-aware SSL tasks. Therefore, investigating a more sophisticated multi-task pretraining scheme to fully incorporate the power of KBA and context-aware SSLs is a good future research direction. We include further discussion in the Appendix Section ``Additive Effects of SSL Tasks'' to demonstrate how each individual SSL task contributes.    


\begin{table}
  \label{few_shots_table}
  \centering
  \small
  \resizebox{0.49\textwidth}{!}{
  \begin{tabular}{lllllll}
    \toprule
    Data quantity & CSR & USR & GSR & UCSR & UGSR & OSR \\
    \midrule
    40\%  &  \textbf{0.848} & 0.303 & 0.714 & 0.247 & 0.189 & 0.173 \\
    80\%  &  0.846 & \textbf{0.320} & \textbf{0.718} & 0.255 & 0.201 & 0.180 \\
    100\%  & 0.825 & 0.373 & 0.652 & \textbf{0.301} & \textbf{0.224} & \textbf{0.200} \\
    \bottomrule
  \end{tabular}
  }
  \caption{Human evaluation of our KBA+SSL (256) model output using different percentages of our parallel corpus.}
  \label{tab:fewshot}
\end{table}

\subsubsection{Fine-tuning Data Quantity Effects}
We repeat our fine-tuning experiment for KBA+SSL (256) on 40\% and 80\% subsets of our parallel corpus. We find that USR, UCSR, UGSR, and OSR drop compared to using the complete set of parallel sentences. Surprisingly, we found that both CSR and GSR improve when fewer training samples are used. The model might focus less on transferring into target sentence style but more on reconstructing the original sentence in these restricted data conditions. Compared to the three baseline models, both 40\% and 80\% of parallel data mined sentences outperform baseline models on UCSR, UGSR and OSR. This finding demonstrates the importance of the KBA+SSL tasks and mild parallel data dependency of the pipeline.



\subsubsection{Effect of Embedding Size} 
Table \ref{tab:ssl} shows that training a larger LM on KBA+SSL improves CSR and GSR, leading to a 9.6\% relative improvement in OSR. Although USR stays the same, KBA+SSL Large can further enhance effective style transfer, indicating by 6\% relative improvement in UCSR and 5\% in UGSR. Increased size giving better results is consistent with previous work \cite{devlin2019bert, lewis2019bart, DBLP:journals/corr/abs-1908-05378}.

\begin{table}
  \label{auto_results_table}
  \centering
  \small
  \resizebox{0.45\textwidth}{!}{%
  \begin{tabular}{llll}
    \toprule
    Model & Style Acc & BLEU & PPL \\
    \midrule
    Style Transformer \cite{dai2019style} &  0.43 & 61.2 & 207\\
    DeleteAndRetrieve \cite{li-etal-2018-delete} & 0.64 & 30.6 & 141\\
    ControlledGen \cite{hu2018controlled} & 0.14 & \textbf{80.4} & 171 \\ \midrule
    Basic Tr. & \textbf{0.66} & 19.8 & 338 \\
    KBA Pretraining Only &  0.62 & 37.1 & 240\\
    SSL Pretraining Only & 0.41 & 59.2 & 163\\
    SSL + KBA Pretraining & 0.63 & 39.2 & 159\\
    SSL + KBA Large (512) & 0.61 & 40.2 & \textbf{127}\\
    \bottomrule
  \end{tabular}
    }
    \caption{Automatic evaluation results: BLEU score measures content similarity between input and output. Style Acc shows the percentage of transferred sentences that labeled as corresponding style by the pretrained style classifier. PPL indicates the fluency of the generated sentence.}
  \label{tab:autoeval}
\end{table}

\begin{table*}
  \label{sentences_table}
  \centering
  \begin{adjustbox}{width=\textwidth}
  \begin{tabular}{ll}
    \toprule
    \multicolumn{1}{c}{\textbf{Model Name}} & \multicolumn{1}{c}{Expert Input $\rightarrow$ Generated Layman Sentences} \\
    \cmidrule(r){1-1}
    \cmidrule(r){2-2}
    \textit{Expert Input} & Fluid accumulation in the lungs may cause \textbf{dyspnea} and \textbf{crackles on auscultation} .\\
    \cmidrule(r){1-1}
    \cmidrule(r){2-2}
    Style Transformer & Fluid accumulation in the lungs may cause \textcolor{red}{attention} and \textcolor{red}{literally on 4.4} .\\
    \cmidrule(r){1-1}
    \cmidrule(r){2-2}
    DeleteAndRetrieve & Fluid may cause \textcolor{red}{various symptoms ( such as a head injury ) }.\\
    \cmidrule(r){1-1}
    \cmidrule(r){2-2}
    ControlledGen & Fluid accumulation in the lungs may cause dyspnea and crackles on \textcolor{red}{pupils} .\\
    \cmidrule(r){1-1}
    \cmidrule(r){2-2}
    Basic Tr. & Fluid accumulation in the lungs may cause \textcolor{red}{shortness of liquids during the pregnancy }.\\
    \cmidrule(r){1-1}
    \cmidrule(r){2-2}
    KBA+SSL & The fluid accumulation in the lungs may cause \textcolor{red}{difficulty breathing ( dyspnea )} and \textcolor{red}{airway narrowing ( auscultation )} \\
    \cmidrule(r){1-1}
    \cmidrule(r){2-2}
    KBA+SSL (Lg) & Fluid may be surgically in the lungs and may cause \textcolor{red}{shortness of breath} .\\
    \cmidrule(r){1-1}
    \cmidrule(r){2-2}
    \textit{Gold Reference} & If fluid accumulates in the lungs , people may become \textbf{short of breath} .\\
    \midrule
    \multicolumn{1}{c}{\textbf{Model Name}} & \multicolumn{1}{c}{Layman Input $\rightarrow$ Generated Expert Sentences} \\
    \cmidrule(r){1-1}
    \cmidrule(r){2-2}
    \textit{Layman Input} &  The \textbf{lung infection} may worsen , usually only in people with a \textbf{weakened immune system} . \\ 
    \cmidrule(r){1-1}
    \cmidrule(r){2-2}
    Style Transformer & The lung \textcolor{red}{commercial} may worsen , usually only in \textcolor{red}{patients} with a \textcolor{red}{Adie} immune system .\\
    \cmidrule(r){1-1}
    \cmidrule(r){2-2}
    DeleteAndRetrieve & The lung infection may worsen , usually in only \textcolor{red}{patients} with a weakened immune system .\\
    \cmidrule(r){1-1}
    \cmidrule(r){2-2}
    ControlledGen & The lung infection may worsen , usually only in people with a weakened immune system .\\
    \cmidrule(r){1-1}
    \cmidrule(r){2-2}
    Basic Tr. & Lung \textcolor{red}{damage} is often worse , usually only when \textcolor{red}{patients} are \textcolor{red}{immunocompromised} people .\\
    \cmidrule(r){1-1}
    \cmidrule(r){2-2}
    KBA+SSL & The infection may worsen , usually only in \textcolor{red}{immunocompromised} patients .\\
    \cmidrule(r){1-1}
    \cmidrule(r){2-2}
    KBA+SSL (Lg) & \textcolor{red}{Pulmonary} infection may worsen , usually only in \textcolor{red}{patients with impaired immunocompromise }.\\
    \cmidrule(r){1-1}
    \cmidrule(r){2-2}
    \textit{Gold Reference} & Extensive \textbf{pulmonary} involvement is uncommon in otherwise healthy people and occurs mainly in those who are \textbf{immunocompromised} .\\
    \bottomrule
  \end{tabular}
  \end{adjustbox}
  \caption{Examples of baseline and our model outputs. Red words are model's modification from the input sentence. Bold black words are expected modifications on medical concepts.}
  \label{tab:casestudy}
\end{table*}

\subsection{Correlation to the Automatic Evaluation} 
To investigate the quality of autmoated metrics we compute a system-level correlation between BLEU score and human judgement CSR, between style accuracy and our human judgement USR and between PPL and human judgement GSR. Similar to previous results \cite{li-etal-2018-delete, cao-etal-2020-expertise}, we find that the BLEU score has moderate correlation to human evaluation CSR with Pearson correlation (PCC) $0.64$ ($p=0.086$). However, BLEU score is not a reliable indicator in expert layman style transfer, as it tends to penalize the semantically-correct phrases when they differ from the surface form of the reference \cite{DBLP:journals/corr/abs-1904-09675}. We find that Style Accuracy has the moderate correlation to human evaluation USR with PCC $0.46$ ($p=0.257$). Similar to previous finding \cite{cao-etal-2020-expertise}, we observed that Style Classifier can be easily fooled and achieving high accuracy by adding random target style words, e.g. ``patient'', into layman sentences and ``people'' into expert sentences. However, none of those transforms is valid because they don't improve layman or expert's understandings of the original sentences. Although, our KBA+SSL Large performs the best under PPL evaluation, we find a weak correlation between PPL and our human judgement GSR, with PCC $-0.38$ ($p=0.352$). Overall, we conclude that three automatic evaluation metrics can be useful for model developments as they exist some correlation to the human evaluation. However, human evaluation is non-replaceable at the current stage. We include further discussion and one concrete example in the Appendix Section ``Case Study of Automatic Evaluation'' and Table \ref{tab:evalcasestudy}. 

\subsection{Case Study}
In the first example (E2L) of Table 4, both Style Transformer (Style Tr) and ControlledGen make lexical substitutions to the target style words. However, those changes cause complete deviation from the sentence meanings. In most cases, ControlledGen stays the same as input, which is the reason that it achieves the high CSR and GSR. DeleteAndRetrieve and Basic Tr. are the most progressive baseline models in changing sentence style which seems to be the reason why both of them achieve competitive USRs. However, neither of them achieve effective style transfer, as the transferred sentences are both disfluent and deviating from original meanings, indicated by their low CSR, GSR, UCSR and UGSR. In our two best performing systems, KBA+SSL and KBA+SSL Large, both models are able to accurately translate ``dyspnea'' to either ``difficulty breathing'' or ``shortness of breath.'' Moreover, our KBA+SSL is able to deduce the reason of ``crackles on auscultation'' as ``airway narrowing''.

In the second example (L2E), ControlledGen copies input to the output. Style Transformer replaces medical vocabulary with random target style words. DeleteAndRetrieve replaces ``people'' to ``patients''. Both Basic Tr. and KBA+SSL are correctly able to translate ``weakened immune system'' to ``immunocompromised.'' However, ``lung damage'' and ``infection'' slightly change the meaning. In the end, only our KBA+SSL Large is able to translate all medical words correctly into the target styles while retaining the sentence meaning (See Appendix Table \ref{tab:fullcasestudy} for more examples).

\section{Related Work}
\paragraph{Text Style Transfer} Due to the limited availability of parallel corpora, most prior work relies unsupervised learning. One approach disentangles style and content representations to generate target-style text sequences by directly manipulating latent representations \cite{shen2017style, hu2018controlled, john-etal-2019-disentangled}.
Another approach synthesizes parallel expert-layman sentence pairs through back translation \cite{prabhumoye2018style, DBLP:journals/corr/abs-1808-07894, lample2018multipleattribute} or cyclic reconstruction \cite{dai2019style, DBLP:journals/corr/abs-2010-00735} to enable supervised learning. 

\citet{jin2020imat} iteratively harvest pseudo-parallel sentences for supervised learning, but this small-scale data mining cannot generate adequately large parallel corpora. 
\citet{malmi-etal-2020-unsupervised} replace source words to the target words using two pretrained masked language models. Similar to our KBA task, \citet{li-etal-2018-delete} reconstruct sentences after source-target word replacement using edit distance. However, simple pretraining scheme and inaccurate edit distance measures restrain their applications to our task. As a result, for this defined task \cite{cao-etal-2020-expertise}, domain-specific terminological correspondence knowledge and large parallel corpus generation are our main focuses.

\paragraph{BiText Data Mining} 
Margin-based criterion has demonstrated good performance in low resources setting \cite{chaudhary2019lowresource, koehn-etal-2019-findings}, LASER embedding \cite{DBLP:journals/corr/abs-1812-10464} and margin-based bi-text mining \cite{schwenk2020ccmatrix}. Margin-based embedding has been widely studied in bilingual and multilingual sentence representations \citet{Bouamor2018ParallelSE, gregoire-langlais-2017-bucc}. But, there is limited prior works that applies this idea to monolingual text style transfer and uses semantic embedding to create pseudo parallel supervisions.



\paragraph{Self-Supervised learning} SSL aims to train a neural network with automatically generated data \cite{peters-etal-2018-deep, devlin2019bert, Radford2018ImprovingLU}. 
There are two existing approaches for pretrained language models, feature-based learning \cite{peters-etal-2018-deep} and fine-tuning \cite{devlin2019bert, Radford2018ImprovingLU}. 
To tackle medical domain-specific terminologies, there are many variants \cite{peng2019transfer, alsentzer2019publicly, beltagy2019scibert, Lee_2019}. But, all these works either pretrain on an encoder or a decoder. Therefore, they are not good fits for seq-to-seq generation.
pretrained encoder-decoder framework \cite{song2019mass, lewis2019bart} has been successful in many downstream sequence generation tasks like text summary, machine translation and question answering. SSL on text style transfer remains under-studied. 


\section{Conclusion}
We built a large-scale parallel corpus extending the MSD dataset using margin-based criterion. We introduced a novel pretraining task, knowledge base assimilation, which combined with established SSL tasks produces a high-quality LM to fine-tune with the parallel corpus. 
This model outperforms unsupervised baselines considerably on human evaluations. We hope that future work will explore a more sophisticated pretraining scheme to fully incorporate KBA and context-aware SSLs and assimilate more complex knowledge graph structures into LMs by extending KBA to novel domains.

\section{Ethics Statement}
As we restricted our training to the publicly available descriptive data, systems trained on it have no risk of personal information leaks beyond those generally associated with any web search. The biggest ethical risk to this work is that erroneous transfers could be misleading to users leading to negative clinical outcomes, if and when used in applications. Extensive human evaluations on the system outputs are required at the current stage before pushing this model into the clinical applications.


\section*{Acknowledgements}
This work was supported in part by the National Science Foundation Graduate Research Fellowship under Grant No.1650114 and by the National Science Foundation award No.2048122. We would also like to thank the Robert N. Noyce Trust for their generous gift to the University of California via the Noyce Initiative. The views and conclusions contained in this document are those of the authors and should not be interpreted as representing the sponsors

\bibliography{anthology,custom}
\quad\newpage
\appendix
\section{Appendix}
\label{sec:appendix}

\subsubsection{Additive Effects of SSL Tasks}
In Table \ref{tab:add_ssl}, we demonstrate the individual importance of each SSL task and how each task is complimentary to the other tasks by linearly adding one task at a time to the pretraining scheme. To do this we pretrain a progression of models where tasks are added in order of increasing complexity.  We observe a general trend that models with more SSL tasks converge faster during fine-tuning and OSR increases as more SSL tasks are added. Furthermore, the results demonstrate that each task is making solid contributions, driving the overall performance higher. 

By pretraining KBA on our transformer model, it intuitively improves USR by learning of terminological mappings between two styles. Surprisingly, as the model learns to reconstruct sentences from the replaced target style medical words, both content and grammar scores improve. Those improvements across all three criteria enhance our model's faculty for \textit{Effective Style Transfer}, indicated by improvement of all UCSR, UGSR and OSR. With the \textit{Mask} added to \textit{KBA} during pretraining, both USR, CSR, UCSR and UGSR are improved, leading to improvement in the OSR. This performance validates our intuition that context-aware learning of medical concepts can improve the model's content understanding and further benefit text style transfer by enhancing effective style transfer. On the contrary, adding the \textit{Switch} task into the pretraining scheme leads to a boost in both content and grammar scores but a decrease in the USR. This behavior confirms our intuition that the Switch task can improve the syntactic awareness of the generated text through word order learning. As this learning scheme improves our \textit{Effective Style transfer} in UGSR, overall performance OSR is improved. With the \textit{Delete} task added in, the model demonstrates a strong improvement in USR. Unlike previous variants, it improves CSR, USR and GSR at the same time. The improvement of USR is effective as both UCSR and UGSR are improved. The success of the \textit{Delete} task verifies our assumption that compared to the \textit{Mask} task, it can gain deeper understandings since contextual and positional information can be learned concurrently, leading to an overall boost in OSR. 

\subsubsection{Case Study of Automatic Evaluation}In Table \ref{tab:evalcasestudy}, we include one output from our baseline model Style Transformer \cite{dai2019style} and one from our KBA + SSL Large (512). Based on our human evaluation CSR, we can clearly see that our model preserves the semantic content and Style Transformer deviates from the original meaning. However, due to the surface form evaluation from the BLEU score, Style Transformer is given the  high BLEU score 51.4 and our KBA + SSL Large (512) and golden reference sentence only receive 0. The pretrained style classifier predicts both Style Transformer's output and our output as layman style sentences. However, human rating only gives 0.33 to the Style Transformer's output, as its transform is surface-level layman word replacements without improving reader's understanding over the original text. PPL score well distinguishes our output from Style Transformer's output by giving 49.2 vs 262 and it matches to our human rating 1 vs 0. However, our generated text only has surface differences compared to the golden reference sentence but the PPL gives much worse score 146 to the golden reference sentence. Therefore, PPL is also not a reliable indicator. This case study further suggests that human evaluation is essential to assess the quality of the generated text in the current stage.

\subsubsection{Implementation of Human Evaluation Interface} 
\begin{figure*}
    \centering
    \includegraphics[scale=0.70]{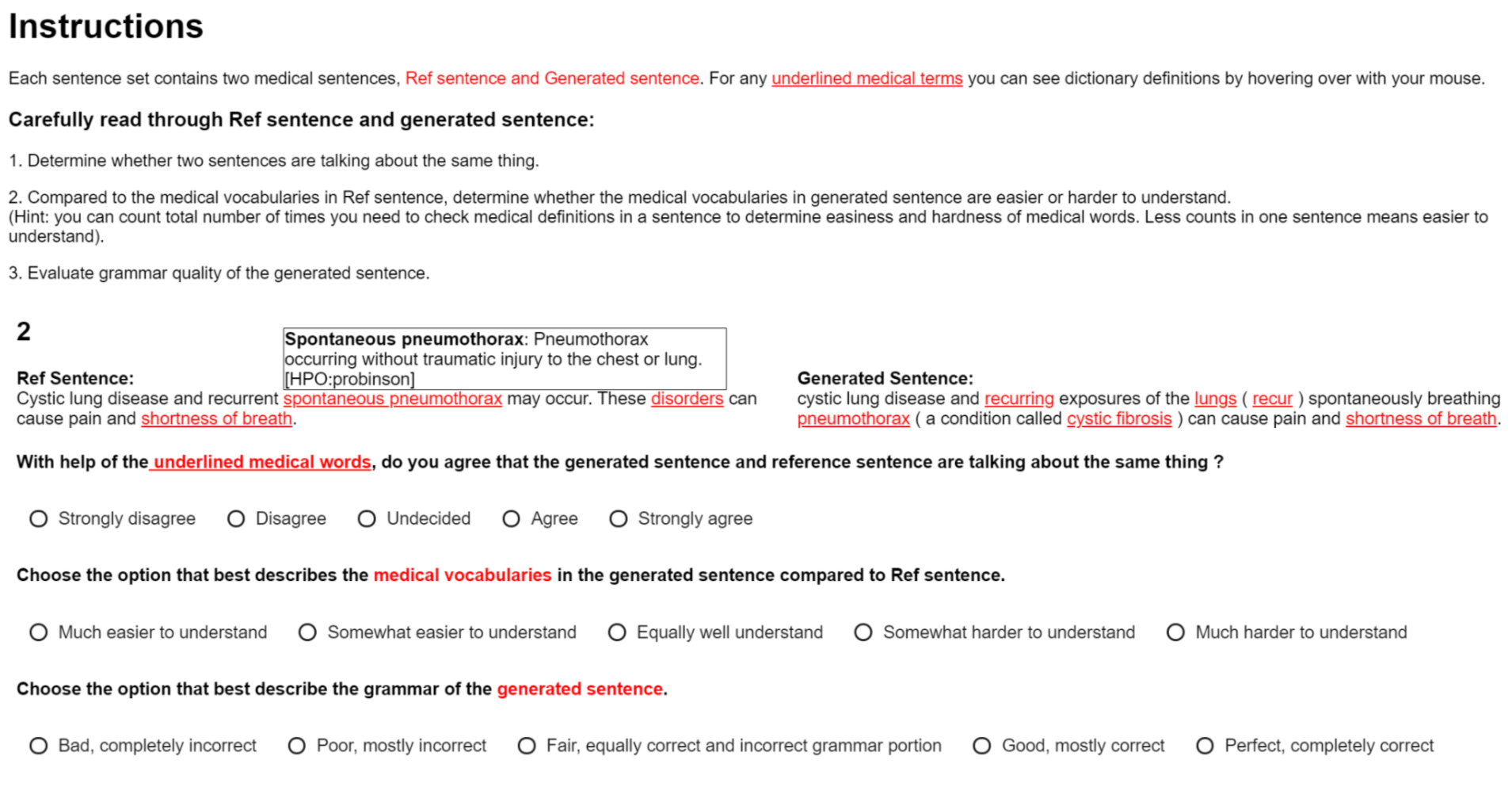}
    \caption{A screenshot of the Human Evaluation Interface with medical terminology definition popups.}
    \label{fig:mturk}
\end{figure*}

To ensure the layperson crowdworkers understand both sentences, including the complex medical terminology they might contain, we use the UMLS system maintained by the US National Institutes of Health (NIH). For each source sentence, a set of `CUI' codes are provided linking medical terms in the source sentence to entries in the UMLS. We use a simple set of heuristics to pull terminology connected to potentially challenging-to-understand words in both the source and translated sentences, and provide the dictionary definitions on mouseover as pop-ups to the crowdworkers. That way, they can use the dictionary definitions to easily determine in real time if two sentences match (see Figure \ref{fig:mturk}).

\begin{table}
  \label{additive_table}
  \centering
  \small
  \resizebox{0.49\textwidth}{!}{%
  \begin{tabular}{lllllll}
    \toprule
    \multicolumn{1}{c}{} & \multicolumn{6}{c}{MSD} \\
    \cmidrule(r){2-7}
    Model & CSR & USR & GSR & UCSR & UGSR & OSR \\
    \midrule
    Style Tr \cite{dai2019style} & 0.703 & 0.281 & 0.615 & 0.176 & 0.058 & 0.113 \\
    DAR \cite{li-etal-2018-delete} & 0.695 & 0.321 & 0.472 & 0.231 & 0.148 & 0.123 \\
    CtrlGen \cite{hu2018controlled} & 0.852 & 0.195 & \textbf{0.739} & 0.103 & 0.147 & 0.086 \\
     \midrule
    Basic Tr. & 0.683 & 0.301 & 0.465 & 0.161 & 0.088 & 0.068 \\
    KBA Pretraining Only & 0.733 & 0.315 & 0.600 & 0.194 & 0.131 & 0.113\\
    KBA + 1 SSL (\textit{Mask}) & 0.774 & 0.318 & 0.540 & 0.238 & 0.148 & 0.125\\
    KBA + 2 SSL (\textit{M.}, \textit{Switch}) & 0.820 & 0.300 & 0.600 & 0.223 & 0.166 & 0.149 \\
    KBA + 3 SSL (\textit{M.}, \textit{S.}, \textit{Delete}) & 0.825 & \textbf{0.373} & 0.652 & 0.301 & 0.224 & 0.200 \\
    KBA + 3 SSL Large (512) & 0.860 & \textbf{0.373} & 0.666 & \textbf{0.320} & \textbf{0.235} & \textbf{0.216} \\
    \bottomrule
  \end{tabular}
    }
    \caption{
    Change in human evaluation results as SSL tasks are progressively added. Baselines, Basic Tr., KBA Pretraining Only, KBA + 3 SSL, and KBA + 3 SSL Large are reproduced from Table 2. Additional rows show progressive gains in performance as SSL tasks (\textit{Mask}, \textit{Switch}, and \textit{Delete}) are added.}
  \label{tab:add_ssl}
\end{table}

\begin{table*}
  \label{eval_sentences_table}
  \centering
  \begin{adjustbox}{width=\textwidth}
  \begin{tabular}{llllllll}
  \toprule
   \multicolumn{1}{c}{\textbf{Model Name}} & \multicolumn{1}{c}{Text} & \multicolumn{1}{c}{BLEU(100)} & \multicolumn{1}{c}{CSR(1)} & \multicolumn{1}{c}{Stye Acc(1)} & \multicolumn{1}{c}{USR(1)} & 
   \multicolumn{1}{c}{PPL} &
   \multicolumn{1}{c}{GSR(1)} \\
   \cmidrule(r){1-1}
   \cmidrule(r){2-8}
   Expert Input & fluid accumulation in the lungs may cause dyspnea and crackles on auscultation . & - & - & - & - &  - & -\\
   Style Transformer & Fluid accumulation in the lungs may cause attention and literally on 4.4 . & 51.4 & 0.333 & 1 & 0.333 & 262 & 0\\
   KBA + SSL Large (512) & Fluid may be surgically in the lungs and may cause shortness of breath . & 0 & 1 & 1 & 1 & 49.2 &  1 \\
   Golden Reference &  If fluid accumulates in the lungs , people may become short of breath . & 0 & - & 1 & - & 146 & -\\
  \bottomrule
  \end{tabular}
  \end{adjustbox}
  \caption{Case Study to compare automatic evaluation and human evaluation results}
  \label{tab:evalcasestudy}
\end{table*}

\begin{table*}
  \label{sentences_table}
  \centering
  \begin{adjustbox}{width=\textwidth}
  \begin{tabular}{ll}
    \toprule
    \multicolumn{1}{c}{\textbf{Model Name}} & \multicolumn{1}{c}{Layman Input Sentences and Generated Expert Sentences} \\
    \cmidrule(r){1-1}
    \cmidrule(r){2-2}
    Layman Input & The most common symptoms , regardless of the type of fluid in the \textbf{pleural space} or its cause , are \textbf{shortness of breath} and chest pain .\\
    \cmidrule(r){1-1}
    \cmidrule(r){2-2}
    Style Transformer & The most common symptoms , regardless of the type of fluid in the pleural space or its cause , are \textcolor{red}{neonatal} of breath and chest pain . \\
    \cmidrule(r){1-1}
    \cmidrule(r){2-2}
    DeleteAndRetrieve & The most common symptoms , regardless of the \textcolor{red}{severity of the valve or cause , is no based on the patient 's symptoms and signs ( see table : some symptoms are reversible ) .}\\
    \cmidrule(r){1-1}
    \cmidrule(r){2-2}
    ControlledGen & The most common symptoms , regardless of the type of fluid in the pleural space or its cause , are shortness of breath and chest pain .\\
    \cmidrule(r){1-1}
    \cmidrule(r){2-2}
    Basic Tr. & Other the most common symptoms , regardless of the \textcolor{red}{subpleen} in \textcolor{red}{the functional reproductive} causes , \textcolor{red}{refer to the left cause is }, and \textcolor{red}{dyspnea} .\\
    \cmidrule(r){1-1}
    \cmidrule(r){2-2}
    KBA + SSL Pretraining & The most common symptoms , regardless of the \textcolor{red}{pleural effusion} , are \textcolor{red}{dyspnea} and chest pain .\\
     \cmidrule(r){1-1}
     \cmidrule(r){2-2}
    KBA + SSL Large (512) & he most common symptoms , regardless of the type of fluid in the pleural space or its cause , are \textcolor{red}{dyspnea} and chest pain . \\
    \cmidrule(r){1-1}
    \cmidrule(r){2-2}
     Golden Ref & Many cause \textbf{dyspnea} , \textbf{pleuritic} chest pain , or both .\\
    \cmidrule(r){1-1}
    \cmidrule(r){2-2}
    Layman Input &  The \textbf{lung infection} may worsen , usually only in people with a \textbf{weakened immune system} . \\
    \cmidrule(r){1-1}
    \cmidrule(r){2-2}
    Style Transformer & The lung \textcolor{red}{commercial} may worsen , usually only in \textcolor{red}{patients} with a \textcolor{red}{Adie} immune system .\\ 
    \cmidrule(r){1-1}
    \cmidrule(r){2-2}
    DeleteAndRetrieve & The lung infection may worsen , usually in only \textcolor{red}{patients} with a weakened immune system .\\
    \cmidrule(r){1-1}
    \cmidrule(r){2-2}
    ControlledGen & The lung infection may worsen , usually only in people with a weakened immune system .\\
    \cmidrule(r){1-1}
    \cmidrule(r){2-2}
    Basic Tr. & Lung \textcolor{red}{damage} is often worse , usually only when \textcolor{red}{patients} are \textcolor{red}{immunocompromised} people .\\
    \cmidrule(r){1-1}
    \cmidrule(r){2-2}
    KBA + SSL Pretraining & The infection may worsen , usually only in \textcolor{red}{immunocompromised} patients .\\
    \cmidrule(r){1-1}
    \cmidrule(r){2-2}
    KBA + SSL Large (512) & \textcolor{red}{Pulmonary} infection may worsen , usually only in \textcolor{red}{patients with impaired immunocompromise }.\\
    \cmidrule(r){1-1}
    \cmidrule(r){2-2}
    Golden Reference & Extensive \textbf{pulmonary} involvement is uncommon in otherwise healthy people and occurs mainly in those who are \textbf{immunocompromised} .\\
    \cmidrule(r){1-1}
    \cmidrule(r){2-2}
    \multicolumn{1}{c}{\textbf{Model Name}} & \multicolumn{1}{c}{Expert Input Sentences and Generated Layman Sentences} \\
    \cmidrule(r){1-1}
    \cmidrule(r){2-2}
    Expert Input & \textbf{pulmonary arteries} are affected , sometimes causing \textbf{pulmonary hypertension} . \\ 
    \cmidrule(r){1-1}
    \cmidrule(r){2-2}
    Style Transformer & Pulmonary arteries are affected , sometimes causing \textcolor{red}{language effectiveness} .\\
    \cmidrule(r){1-1}
    \cmidrule(r){2-2}
    DeleteAndRetrieve & Pulmonary arteries are affected , sometimes causing pulmonary hypertension .\\
    \cmidrule(r){1-1}
    \cmidrule(r){2-2}
    ControlledGen & Pulmonary arteries are affected , sometimes causing pulmonary hypertension .\\
    \cmidrule(r){1-1}
    \cmidrule(r){2-2}
    Basic Tr. & The pulmonary arteries are affected , sometimes causing pulmonary hypertension .\\
    \cmidrule(r){1-1}
    \cmidrule(r){2-2}
    KBA + SSL Pretraining & The pulmonary arteries are affected , sometimes causing \textcolor{red}{high blood pressure} .\\
    \cmidrule(r){1-1}
    \cmidrule(r){2-2}
    KBA + SSL Large (512) & The pulmonary arteries are also affected , sometimes causing the \textcolor{red}{person to clot} . \\
    \cmidrule(r){1-1}
    \cmidrule(r){2-2}
    Golden Reference & Lungs : \textbf{Blood pressure} in the lungs becomes very \textbf{high} ( pulmonary hypertension ) .\\
    \cmidrule(r){1-1}
    \cmidrule(r){2-2}
    Expert Input & fluid accumulation in the lungs may cause \textbf{dyspnea} and \textbf{crackles on auscultation} .\\
    \cmidrule(r){1-1}
    \cmidrule(r){2-2}
    Style Transformer & Fluid accumulation in the lungs may cause \textcolor{red}{attention} and \textcolor{red}{literally on 4.4} .\\
    \cmidrule(r){1-1}
    \cmidrule(r){2-2}
    DeleteAndRetrieve & Fluid may cause \textcolor{red}{various symptoms ( such as a head injury ) }.\\
    \cmidrule(r){1-1}
    \cmidrule(r){2-2}
    ControlledGen & Fluid accumulation in the lungs may cause dyspnea and crackles on \textcolor{red}{pupils} .\\
    \cmidrule(r){1-1}
    \cmidrule(r){2-2}
    Basic Tr. & Fluid accumulation in the lungs may cause \textcolor{red}{shortness of liquids during the pregnancy }.\\
    \cmidrule(r){1-1}
    \cmidrule(r){2-2}
    KBA + SSL Pretraining & The fluid accumulation in the lungs may cause \textcolor{red}{difficulty breathing ( dyspnea )} and \textcolor{red}{airway narrowing ( auscultation )} \\
    \cmidrule(r){1-1}
    \cmidrule(r){2-2}
    KBA + SSL Large (512) & Fluid may be surgically in the lungs and may cause \textcolor{red}{shortness of breath} .\\
    \cmidrule(r){1-1}
    \cmidrule(r){2-2}
    Golden Reference & If fluid accumulates in the lungs , people may become \textbf{short of breath} .\\
    \cmidrule(r){1-1}
    \cmidrule(r){2-2}
    \bottomrule
  \end{tabular}
  \end{adjustbox}

  \caption{Examples of baseline and our model outputs. Red words are model’s modification from the input sentence. Bold black words are expected modifications on medical concepts.}
  \label{tab:fullcasestudy}
\end{table*}


\end{document}